# PMU measurements based short-term voltage stability assessment of power systems via deep transfer learning

Yang Li, *Senior Member*, *IEEE*, Shitu Zhang, Yuanzheng Li, *Senior Member*, *IEEE*, Jiting Cao and Shuyue Jia, *Student Member*, *IEEE*

*Abstract*—Deep learning has emerged as an effective solution for addressing the challenges of short-term voltage stability assessment (STVSA) in power systems. However, existing deep learning-based STVSA approaches face limitations in adapting to topological changes, sample labeling, and handling small datasets. To overcome these challenges, this paper proposes a novel phasor measurement unit (PMU) measurements-based STVSA method by using deep transfer learning. The method leverages the real-time dynamic information captured by PMUs to create an initial dataset. It employs temporal ensembling for sample labeling and utilizes least squares generative adversarial networks (LSGAN) for data augmentation, enabling effective deep learning on small-scale datasets. Additionally, the method enhances adaptability to topological changes by exploring connections between different faults. Experimental results on the IEEE 39-bus test system demonstrate that the proposed method improves model evaluation accuracy by approximately 20% through transfer learning, exhibiting strong adaptability to topological changes. Leveraging the self-attention mechanism of the Transformer model, this approach offers significant advantages over shallow learning methods and other deep learning-based approaches.

*Index Terms*— PMU measurements, short-term voltage stability assessment, power system stability, deep transfer learning, least squares generative adversarial networks, temporal ensembling, Transformer.

## I. Introduction

THE short-term voltage stability (STVS) of a power system, also known as transient voltage stability, refers to the responsiveness of rapidly reacting load components, such as induction motors, and electronically-controlled loads, over a span of a few seconds [1, 2]. In the power system, the increasing prevalence of induction motor loads [3] and the expanding integration of renewable energy sources like solar and wind energy [4] have heightened the issue of STVS.

In recent years, with the widespread use of Wide Area Measurement Systems (WAMS) in power systems, the ability to capture dynamic processes in power systems has been greatly enhanced [5]. Compared to the measurement information of conventional supervisory control and data acquisition (SCADA) systems, phasor measurement unit (PMU) measurements provided by WAMS have the advantage of a high sampling frequency and can measure phase angle [6-8], which provides a new perspective for data-driven short-term voltage stability assessment (STVSA). Our paper focuses on STVSA in interconnected power systems that are involved in the generation, transmission, and distribution of electric power, emphasizing the use of phasor measurement units (PMUs) and considering the influence of dynamic load components, and various operating conditions. Reference [9] mentioned the problem of data loss caused by PMUs and reference [10] showed a method of using PMUs' synchronous measurement data to determine the uncertainty bounds of transmission lines. Reference [11] designed a PMU-based robust state estimation method for real-time monitoring of a power system under different operational conditions, and reference [12] constructed a Kalman filter approach to power system state estimation based on PMUs. The real-time assessment of STVS is based on the dynamic information collected by PMUs after the power system experiences disturbances, and the power system will trigger emergency control in a timely manner when an unstable state is predicted. Therefore, developing a fast and accurate STVSA model for power systems is of great significance.

At present, some pioneering works, such as power system stability assessment methods based on energy function [13] and P-V plane [14], have been carried out to solve STVSA problem. Due to the complexity of the load dynamic driving force of the above methods, it is difficult to analyze and construct a unified STVSA criterion for various power systems. In recent years, some works apply artificial intelligence (AI) in instrumentation and measurement [15]. Reference [16] proposed a method for assessing the voltage stability margin of power systems based on artificial neural networks (ANN). Additionally, reference [17] introduced a method for voltage stability monitoring and assessment based on PMUs and ANN. Reference [18] developed an online evaluation method for short-term voltage

This work is supported by the Natural Science Foundation of Jilin Province, China under Grant YDZJ202101ZYTS149, and Open Project of Key Laboratory of Modern Power System Simulation and Control and Renewable Energy Technology, Ministry of Education, Northeast Electric Power University under Grant MPSS2022-04. (*Corresponding author: Yuanzheng Li.*)

Yang Li, and Shitu Zhang are with the School of Electrical Engineering, Northeast Electric Power University, Jilin 132012, China (e-mail: liyang@neepu.edu.cn; 243984790@qq.com).

Yuanzheng Li is with the School of Artificial Intelligence and Automation, Huazhong University of Science and Technology, Wuhan 430074, China (e-mail: Yuanzheng_Li@hust.edu.cn)

Jiting Cao is with State Grid Chengde Power Supply Company, Chengde 067000, China (e-mail: 1277230823@qq.com).

Shuyue Jia is with the Department of Electrical and Computer Engineering, Boston University, Boston, MA 02215, USA (e-mail: shuyuej@ieee.org).

stability in power systems based on a decision tree (DT), while reference [19] enhanced the understanding of modern distribution-transmission interactions related to STVS. Reference [20] proposed a hierarchical adaptive data analysis method based on extreme learning machine (ELM), which can be employed for real-time analysis in power system STVSA. As STVS is intrinsically linked to load dynamics, reference [21] introduced a data-driven method based on support vector machine (SVM) to modify the load dynamic stability index. Reference [22] constructed an integrated model composed of ELM and random vector functional link network (RVFL). Compared to employing ELM and RVFL independently, this integrated model boasts higher accuracy.

Compared to the shallow networks mentioned above, deep learning techniques are more capable of effectively uncovering the underlying complex relationships among big data. Specifically, they have gained substantial popularity in the realm of instrumentation and measurement due to their prowess in extracting features from voluminous data and making precise predictions. Recently, some deep learning algorithms have started to be employed in the domain of power system security assessment, as it can extract the potential patterns from a large volume of system operation data [23]. Reference [24] introduced a deep learning approach, which by learning the time dependencies of system dynamics after a disturbance, established an evaluation model based on long short-term memory (LSTM). Reference [25] developed a data-driven STVSA method based on graph convolution network (GCN). Reference [26] introduced an STVSA model based on graphical neural network (GNN), which can quickly identify fast voltage collapses (FVC) and delayed voltage recovery events caused by faults. Reference [27] proposed a real-time STVSA method by combining temporal convolutional neural network (CNN) and LSTM. Based on spatio-temporal information, reference [28] proposed an online STVSA method by combining GCN and LSTM networks.

Even though deep learning has been introduced into the STVSA domain, making deep learning models perform well on topological change scenarios and small data remains a significant challenging task. Reference [29] developed a transfer learning method designed to apply a pre-trained model to unknown scenarios, enhancing the scalability of the model. Nonetheless, a substantial gap still remains. In order to underscore the research deficiencies our study seeks to address, and to delineate the contributions of our proposed work, we have conducted a comprehensive comparison with state-of-the-art methods and recent studies. This comparison is presented in TABLE I, where the symbols ✓ and ✗ denote whether the corresponding method is adopted or not adopted by the references. In addition, the full names of the corresponding methods are as follows: deep learning (DL), data augmentation (DA), labeling data (LA), statistical measures (SM) and transfer learning (TL).

In TABLE I, the listed items are introduced in detail as follows: DL signifies that the STVSA model is established based on deep learning algorithms; DA indicates that a limited small dataset is augmented using data augmentation techniques; LA denotes the data labeling algorithms are used; TL represents the transfer learning algorithms are employed; SM stands for statistical measures, such as operating characteristic curve, Matthews correlation coefficient, and F1-score, which are used to comprehensively measure the overall characteristics of the constructed assessment model.

TABLE I
COMPARISON OF THE PROPOSED APPROACH WITH RELATED WORKS

| References | Items | | | | |
|---|---|---|---|---|---|
| | DL | DA | LA | SM | TL |
| [21] | × | × | × | × | × |
| [22] | × | × | × | × | × |
| [23] | √ | × | × | × | × |
| [24] | √ | × | × | × | × |
| [25] | √ | × | × | √ | √ |
| [26] | √ | × | × | × | × |
| [27] | √ | × | × | × | √ |
| Proposed approach | √ | √ | √ | √ | √ |

Unfortunately, once the network topology of the power system alters, the distribution and structure of the dataset will correspondingly change. At the same time, the original assessment model is no longer applicable and the accuracy will be greatly reduced when performing an online assessment. The traditional solution is to retrain the model, but it will take a lot of time. Considering the potential correlations between different faults due to physical factors, this study introduces transfer learning to enhance adaptability to topological changes and assessment performance.

Matrix classification networks such as the multi-class fuzzy support matrix machine [30], the non-parallel bounded support matrix machine [31], and the deep stacked support matrix machine [32] have made notable strides in pattern classifications. Despite these advancements, they continue to grapple with practical limitations, including adaptability to network topology changes and efficient processing of small datasets. To overcome these limitations, this study presents a novel STVSA method based on deep transfer learning, which integrates semi-supervised clustering, data augmentation, and classification using the Transformer [33]. This method exhibits significant advantages in handling sequential data, learning long-distance dependencies, and in the scope of deep and transfer learning. There are four contributions in this paper:

(1) Deep transfer learning is introduced to STVSA for the first time in this paper, deeply mines the correlation between different fault datasets. It can not only verify the adaptability of the model to network topology change and it is a novel approach in the STVSA field.

(2) This work employs temporal ensembling, an advanced semi-supervised clustering method, to address the challenge of sample labeling arising from the lack of STVS criteria. In order to obtain a deep learning model with good performance based on small datasets, least squares generative adversarial networks (LSGAN) is introduced in this paper to generate good synthetic samples.

(3) In the online assessment stage, our work utilizes PMUs

for executing STVSA in real time. This approach leverages the precision of PMUs to swiftly determine assessment outcomes, demonstrating the significant role of advanced instrumentation in enhancing power system stability assessments.

(4) This paper introduced the Transformer model to improve the assessment accuracy in power system, which is a more advanced machine learning method. Compared to other methods, this method can achieve better performance with a shorter observation time window (OTW) length in STVSA.

## II. RELATED ALGORITHMS

### A. Temporal ensembling

Thus far, academia lacks a universally established quantitative criterion for determining STVS, complicating the process of obtaining precise labels for samples within the dataset necessary for the construction of deep learning models. If the labels are assigned individually in accordance with engineering criteria, it would result in a conservative outcome, as well as a tedious and time-consuming handling process. According to the common sense of voltage stability, it's known that: if all bus voltages remain above 0.9 p.u. after a disturbance, the power system can be considered stable; conversely, if the voltage at all buses in the power system falls below 0.7 p.u. without recovery, it can be classified as an unstable system [34]. A limited number of samples can be assigned distinct labels, which in turn can guide the clustering process for the larger collection of unlabeled samples. Utilizing these labeled samples, the semi-supervised clustering method can be used to determine the class label of each data sample. At present, we utilize an innovative semi-supervised clustering method known as temporal ensembling.

Temporal ensembling, which improves upon the π-model, incorporates temporal elements while adhering to the fundamental principles of consistency regularization. Within the scope of classification problems, the standard cross-entropy loss is calculated for labeled data, and consistency loss is calculated for all data, including both labeled and unlabeled. The weights of these two elements are combined to calculate the overall loss, which is defined as [35]:

$$\text{loss} \leftarrow -\frac{1}{|B|}\sum_{i\in(B\cap L)}\log z_i[y_i] \\ + w(t)\frac{1}{C|B|}\sum_{i\in B}\|z_i - \tilde{z}_i\|^2. \quad (1)$$

where, $L$ is the labeled dataset in the entire training dataset. The first line represents the cross-entropy loss, where $B\cap L$ implies that only labeled dataset is considered. The second line represents the consistency loss, which directly calculates $L$ distance $\|z_i - \tilde{z}_i\|^2$ between $B\cap L$ and $\tilde{z}_i$. It measures the sum of the distances of the classification results for each class to make them as close as possible. The consistency loss of all samples is involved in the calculation and the $w(t)$ is a weight function which changes over time and is used to control the weight loss of consistency. Since the network parameters are initially random, the data may not be well-fitted, causing the semi-supervised part's loss to be considerably large. If $w(t)$ is too high, it can adversely affect the network's training. Hence, $w(t)$ increases gradually over time.

A notable drawback of the π-model is that it necessitates two forward propagations for each input to compute the consistency loss, which may be inefficient. On the other hand, temporal ensembling considers generating a standard $\tilde{z}$ for comparison during training, and each time the output $z_i$ is consistent with the annotation $\tilde{z}$. The consistency loss is calculated as $\sum\|z_i - \tilde{z}\|^2$.

The value of $\tilde{z}$ is continuously $z_i$ updated by each output $z$ using the Exponentially Moving Average (EMA), and the output of the previous epochs model is averaged through EMA, which also secretly utilizes the idea of ensembling learning. The updating is governed by the following equations:

$$Z = \alpha Z + (1-\alpha)z. \quad (2)$$
$$\tilde{z} = Z/(1-\alpha^t). \quad (3)$$

Both $z$ and $\tilde{z}$ are required to be initialized to 0, while (1-$a^t$) is used to ensure that $\tilde{z}$ and $z$ are always on the same scale. Upon implementing temporal ensembling, each sample is passed through the network just once; and similarly, each input results in a single evaluation of the network. This efficiency accelerates the training process.

### B. Least Squares Generative Adversarial Network

Generative adversarial network (GAN) is an emerging powerful tool for data preprocessing, and its associated methods have been employed to address the uncertainties inherent in renewable energy outputs in complex dispatch challenges [36, 37]. The task of the generator $G$ is to generate samples that are as realistic as possible, while the discriminator $D$ aims to distinguish between the generated samples and real samples with the highest accuracy. The objective function of GAN is shown below [38]:

$$\min_G \max_D V(D,G) = \mathbb{E}_{x\sim\mathbb{P}_r(x)}[\log D(x)] + \mathbb{E}_{z\sim\mathbb{P}_z(z)}[\log(1-D(G(z)))]. \quad (4)$$

In this model, $\mathbb{P}_r(x)$ is a probability distribution obeyed by real data $x$, $\mathbb{P}_z(z)$ is the probability distribution followed by noise $z$ and $\mathbb{E}_{z\sim\mathbb{P}_z(z)}$ is the expected value.

Facing the issue of vanishing gradients, traditional GAN employs cross-entropy loss as a mitigating measure. In contrast, LSGAN offers a more effective solution by leveraging the least squares function with binary encoding as the loss function. The objective function of LSGAN is formulated by [39]:

$$\min_D L(D) = \frac{1}{2}E_{x\sim P_{data}(x)}[(D(x)-1)^2] + \frac{1}{2}E_{Z\sim P_Z(Z)}[D(G(Z))^2]. \quad (5)$$

$$\min_G L(G) = \frac{1}{2}E_{z\sim P_z(z)}[(D(G(z))-1)^2]. \quad (6)$$

In LSGAN, the one-hot label, $y$, is appended to the end of the noise, $z$, and the real data, $x$, respectively [39], serving as new input in the objective function. The objective functions are defined as follows:

$$\min_D L(D) = \frac{1}{2}E_{x,y\sim P_{data}(x,y)}[(D(x,y)-1)^2] \\ + \frac{1}{2}E_{z\sim P_z(z), y\sim P_{data}(x,y)}[(D(G(z,y),y))^2]. \quad (7)$$

$$\min_G L(G) = \frac{1}{2} E_{z \sim P_z(z), y \sim P_{data}(x,y)} \left[ (D(G(z,y),y) - 1)^2 \right]. \quad (8)$$

It's important to underscore that our approach leverages both LSGAN and temporal ensembling to yield optimal results. This integrated method fosters model adaptability and bolsters generalization, serving as an effective solution for addressing small sample learning scenarios. These processes are applied in sequence, underscoring the synergy they bring to our methodology.

*C. Transformer*

The Transformer is a cutting-edge algorithm that has garnered extensive acclaim in the realm of natural language processing. The Transformer fundamentally embodies a sequence-to-sequence architecture typically bifurcated into two critical components: the encoder and the decoder. Within the encoder, multiple identical layers are stacked, each composed of two sub-layers: a multi-head self-attention mechanism and a feed-forward neural network (FNN). Based on the encoder's original multi-head self-attention and FNN, the decoder introduces an additional encoder-decoder attention sub-layer. Moreover, every sub-layer incorporates a residual connection and layer normalization. To consider the order of the model input, the Transformer appends a positional encoding at the input layer.

The specific structure of Transformer is meticulously illustrated in [33]. Self-attention is a variant of attention that processes sequences by replacing each element with a weighted average of the remainder of the sequence, which is given by:

$$\text{Attention}(Q, K, V) = \text{softmax}(\frac{QK^T}{\sqrt{d_k}})V, \quad (9)$$

where Q, K, and V represent the input query, key and value feature matrix, respectively; and $d_k$ is the dimension of matrix K. Reference [33] proposed the concept of multi-head attention, which combines input features from different positions with different weights and splices the results to obtain the final result. The formulas are as follows:

$$\text{MultiHead}(Q, K, V) = \text{Concat}(\text{head}_1, ..., \text{head}_h)W^O, \quad (10)$$

$$\text{head}_i = \text{Attention}(QW_i^Q, KW_i^K, VW_i^V), \quad (11)$$

where, $W^Q$, $W^K$ and $W^V$ are weight matrices, and $W^O$ is an additional matrix. The multi-head attention mechanism enables the model to assimilate information from disparate subspaces at various positions, thereby enhancing its performance.

## III. MODEL STRUCTURE OF TRANSFER LEARNING-TRANSFORMER

Aiming to fuse the strengths of both deep transfer learning and the Transformer model for STVSA, we have crafted a model structure termed Transfer Learning-Transformer. This combination is inspired by transfer learning's capacity to extrapolate knowledge across diverse yet related fault scenarios, and the Transformer's adeptness in handling time-series data of variable lengths. Collectively, these could potentially enhance the accuracy and adaptability of STVSA models.

Transfer learning is a machine learning technique where a pre-trained model is adapted for a different but related problem. It allows us to leverage the understanding from one task to solve another similar task, rather than training a new model from scratch. This is particularly beneficial for situations where we have limited data or similar tasks with slight differences. On the other hand, the Transformer model, known for its self-attention mechanism, is particularly well-suited to handle dependencies in sequence data by allowing direct paths between distant input and output positions in the sequence. This capability of the Transformer enables it especially suited to analyzing the dynamic processes of power systems after disturbances, where temporal dependencies could span across variable lengths of time. By fusing the principles of transfer learning and Transformer, we seek to develop a more accurate, robust, and adaptable STVSA model.

The Transfer Learning-Transformer model structure is described in detail in **Fig. 1**.

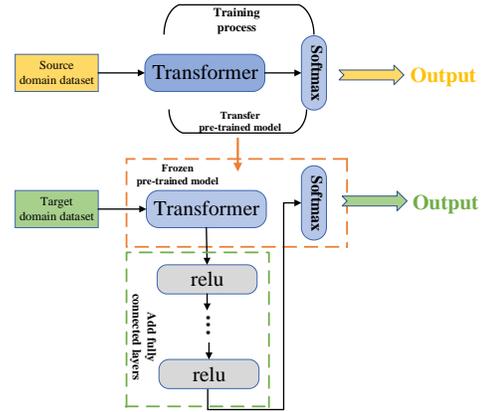

**Fig. 1.** Transfer Learning-Transformer model structure.

Our approach employs Transformer to leverage its self-attention mechanism in extracting meaningful features from PMU measurements. Pre-trained on a source domain dataset that encompasses various power system fault scenarios, the Transformer is adept at uncovering universal features relevant across diverse power systems and faults, forming the cornerstone of our feature extraction strategy.

The extracted features are further honed during the fine-tuning stage using specific data from the target domain dataset, enabling the network to adapt these features to its unique characteristics and fault types. This phase is instrumental in ensuring the model's adaptability as it amalgamates the unique insights of the target domain while preserving the valuable knowledge gleaned from the source domain dataset.

The framework of transfer learning establishes the crucial bridge between the source domain dataset and target domain dataset. By refining the generic features on the target domain dataset, they are aligned with its specific fault types and characteristics, enhancing fault type recognition. By this means, by deeply mining the correlations between different faults and exploiting the relationship between the source and target domain datasets, our proposed method significantly outperforms other assessment approaches in this field.

## IV. PROPOSED STVSA INTELLIGENT SYSTEM

### A. Acquisition of post-fault state information

The STVS depends not only upon the initial operating conditions (OCs) but also upon the severity of the fault, such as fault type and location. Consequently, post-fault dynamic information is highly indicative of the final stability status of the disturbed power system. However, traditional measurement systems like SCADA have inherent limitations that frequently necessitate the utilization of pre-fault static features in existing studies. To overcome this bottleneck, the recent advancement and industrial application of WAMS have emerged, with its structure depicted in **Fig. 2**.

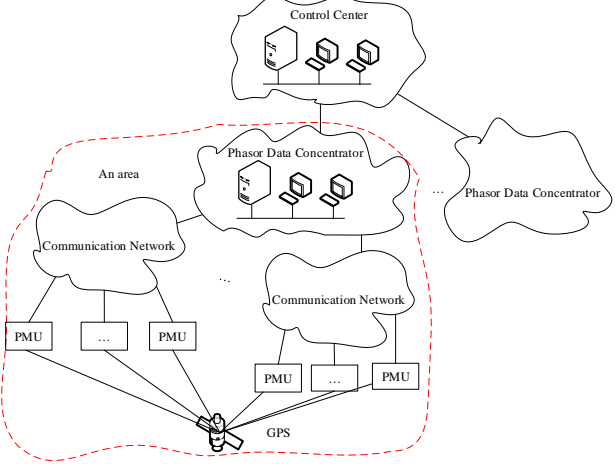

**Fig. 2.** Structure diagram of WAMS.

**Fig. 2** illustrates that WAMS, functioning as a measurement infrastructure spanning multiple areas, typically consists of three core components: PMUs, a communication system, and a control system. Given the ability of WAMS to provide post-fault synchronized measurements, this study places emphasis on harnessing features extracted from PMU data as pivotal predictors for real-time STVSA.

### B. Stability assessment Transformer

Within the structure of the Transformer, the encoder layer is tasked with distilling data features, while the decoder layer capitalizes on these features to make predictions. Given that classification tasks primarily focus on learning data features, we made modifications to the original Transformer. Specifically, we removed the decoder layer and appended fully connected (FC) and Softmax layers after the multi-layer encoder. The simplified structure diagram of Transformer is shown in **Fig. 3**.

The input to the Transformer consists of time-series samples, which are composed of real-time data from the power system measured by PMUs. These samples undergo positional encoding before they are fed into the encoder. As the samples pass through the various layers of the encoder, data features are learned using a self-attention mechanism, and this information is propagated via a feed-forward neural network. Finally, a fully connected layer along with a Softmax layer is appended at the end of the encoder, which outputs the samples used to evaluate the system stability.

### C. STVSA process

In this paper, the proposed STVSA scheme consists of three stages: data processing, offline training, and online application. A detailed illustration is shown in **Fig. 4**.

Data processing stage: This study considers different operating conditions by using time-domain simulations to obtain initial time-series samples, where each time-series sample includes normalized voltage ($U$), active power ($P$), and reactive power ($Q$) at the current sampling moment. Through the annotation of initial unlabeled samples, we leverage temporal ensembling to generate a fully labeled dataset. Subsequently, this dataset is introduced into LSGAN to facilitate data augmentation and is expanded as the final dataset.

Offline training stage: The final dataset is segmented into source and target domain datasets. Both these subsets are further divided into training and testing datasets in a 4:1 ratio, as indicated in [27]. The training set from the source domain is utilized to instruct the stability assessment Transformer. The corresponding testing set serves to evaluate the Transformer's performance, resulting in a pre-trained model. Subsequently, the target domain's training dataset is integrated into this pre-trained model for parameter fine-tuning. Upon meeting the specified assessment indexes, we achieve a fully optimized and trained model.

Online application stage: Once a large disturbance occurs in the system, the input features extracted from PMU measurements will be fed into the trained STVSA model to judge whether the power system is able to maintain stability. If the system is judged to be stable, the condition of the power system will be continuously monitored; otherwise, corresponding control measures will be promptly initiated.

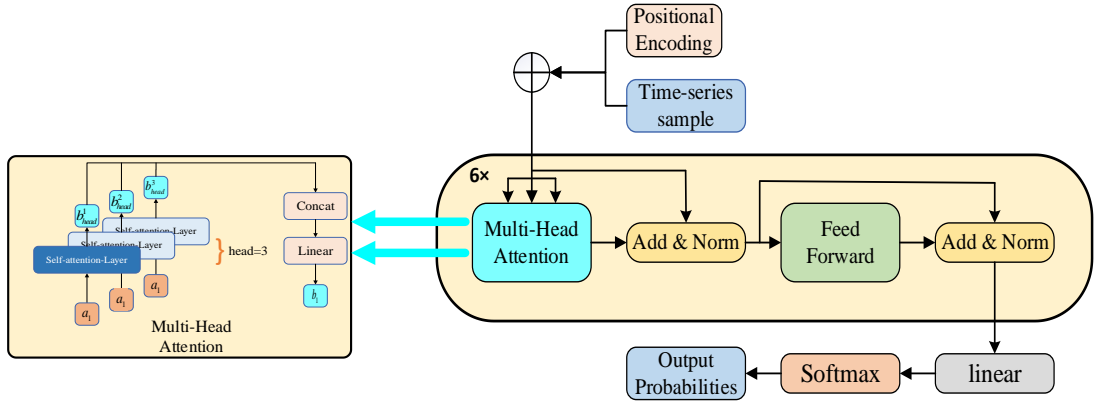

**Fig. 3.** Simplified structure diagram of Transformer.

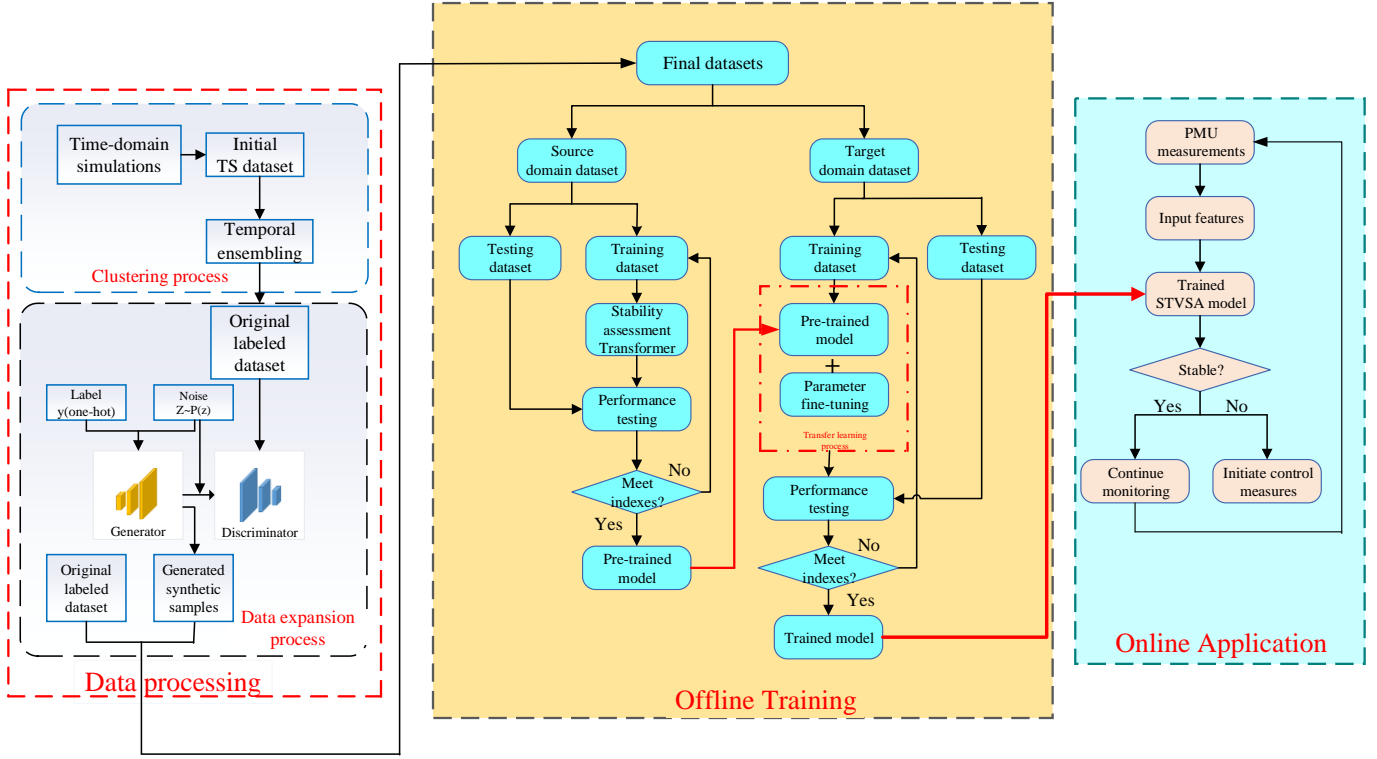

**Fig. 4.** Flow chart of the proposed method.

While accuracy is often employed as a primary evaluation metric in classification models, this paper extends the analytical lens beyond mere accuracy for a more comprehensive performance evaluation. We introduce and utilize other significant metrics, including the area under the curve (AUC), Matthews correlation coefficient (MCC), F1-Score, and silhouette coefficient (SC), all of which play crucial roles in evaluating model performance. These metrics, defined within the context of the confusion matrix, offer a holistic assessment of the model's effectiveness, providing a more nuanced understanding of its capabilities.

In the context of the upcoming formula, we use the following variables: TP (True Positive), which represents instances when stable samples are correctly identified as stable; FP (False Positive), when unstable samples are mistakenly classified as stable; FN (False Negative), when stable samples are incorrectly identified as unstable; and TN (True Negative), when unstable samples are accurately identified as unstable.

1) **Accuracy (ACC)**

In classification evaluation, using ACC is beneficial because it succinctly quantifies the proportion of correctly classified instances relative to the total number of instances. This makes it a straightforward and useful initial indicator of the model's general efficacy. The ACC is shown as [22]:

$$\text{ACC} = \frac{\text{TP} + \text{TN}}{\text{TP} + \text{FP} + \text{FN} + \text{TN}}. \quad (12)$$

2) **Matthews correlation coefficient (MCC)**

The MCC is a robust statistical measure that quantifies the quality of classifications, taking into account all categories of correct and incorrect predictions. It is especially adept at offering a balanced evaluation in cases with highly unbalanced datasets, where it often surpasses metrics that are based solely on accuracy. The MCC is defined as [40]:

$$\text{MCC} = \frac{\text{TP} \times \text{TN} - \text{FP} \times \text{FN}}{\sqrt{(\text{TP} + \text{FP})(\text{TP} + \text{FN})(\text{TN} + \text{FP})(\text{TN} + \text{FN})}}. \quad (13)$$

3) **F1-score**

The F1-score is the harmonic mean of precision and recall, which inherently implies that a high F1-score can only be achieved if both precision and recall are high. If either is low, the F1-score will tend towards the lower value. The F1-score ranges between 0 and 1, with values nearing 1 signifying exceptional performance of the classification model. The specific formula is shown below [24]:

$$\text{Precision} = \frac{\text{TP}}{\text{TP} + \text{FP}}. \quad (14)$$

$$\text{Recall} = \frac{\text{TP}}{\text{TP} + \text{FN}}. \quad (15)$$

$$\text{F1} - \text{score} = 2 \times \frac{\text{Precision} \times \text{Recall}}{\text{Precision} + \text{Recall}}. \quad (16)$$

4) **Area under the curve (AUC)**

The AUC, a widely used performance metric for evaluating classification models, measures the classifier's ability to differentiate between positive and negative classes. Its value ranges from 0.5 to 1, with a higher AUC value indicating superior performance of the classification model.

5) **Silhouette coefficient (SC)**

The SC serves as a valuable indicator for evaluating the quality of clustering results. The SC value is directly correlated with clustering performance, with higher values indicating more cohesive and well-separated clusters. The SC is given by:

$$SC = \frac{1}{N}\sum_{j=1}^{N}\frac{b_j - a_j}{\max(a_j, b_j)}. \quad (17)$$

Let's consider a scenario where the dataset is divided into multiple clusters. In this context, $a_j$ denotes the average distance between a sample and other samples within the same cluster, which quantifies the intra-cluster cohesion; $b_j$ represents the degree of separation between different clusters, indicating the inter-cluster dissimilarity [40].

## V. CASE STUDY

In subsequent experiments, we demonstrate the superior performance of our method compared to other advanced approaches. For these experiments, we harness the well-known IEEE 39-bus test system [40, 41]. Given the challenges associated with obtaining actual operational data from power systems, we resort to a commercial simulation software, PSD-BPA, to generate the initial datasets [40]. All experiments are conducted using PyTorch 1.7.1 and TensorFlow 2.4.0 on Python 3.8. The experimental setup is carried out on a personal computer with the following specifications: Intel Core i5-6300HQ 2.3GHz CPU, 4GB RAM, and GTX 960M GPU.

The main hyper-parameters of LSGAN and Transformer are given in Table II, where *k* denotes the parameter used to control the balance of discriminator and generator. Note that the parameters are chosen by the try-and-error method in this study.

Table II
HYPER-PARAMENTS SETTING OF LSGAN AND TRANSFORMER

| Item | Parameters | Values |
|---|---|---|
| LSGAN | optimizer | Adam |
| | learning rate | 0.0001 |
| | $k$ | 4 |
| | batch size | 32 |
| | epoch | 1000 |
| | number of iterations | 3000 |
| Transformer | number of multi-head attention | 8 |
| | optimizer | Adam |
| | learning rate | 0.0001 |
| | dropout rate | 0.5 |
| | batch size | 64 |
| | epoch | 400 |

### A. Dataset generation

The source domain dataset and the target domain dataset are generated by the PSD-BPA.

1) **Source domain dataset generation**
   a) The load level is set to 80%, 90%, 100%, 110%, and 120% of its base value. To maintain power balance within the system, the output power of the generators is correspondingly adjusted.
   b) The induction motor load ratios are designated to 70%, 80%, 90% of the total load.
   c) The locations for three-phase short-circuit faults are predetermined at 25%, 50%, and 75% of the line length. To account for fault diversity, single-phase short-circuit faults are introduced across all buses.
   d) The transmission line undergoes a short-circuit fault at 0.1s, with fault clearing times designated at 0.15s (near end) and 0.2s (far end).

In accordance with the operating conditions outlined above, a source domain dataset is generated, consisting of 2040 samples. Of these, 979 are stable and 1061 are unstable.

2) **Target domain datasets generation**

Five target domain datasets are created under five kinds of power system network topologies with permanent disconnection of transmission lines. Each distinct disconnection fault topology is depicted in **Fig. 5**, where the red cross marks represent the fault locations in different target domain datasets. The structures of these datasets precisely match the source domain, including identical operating conditions, fault settings, induction motor load ratios, among other factors.

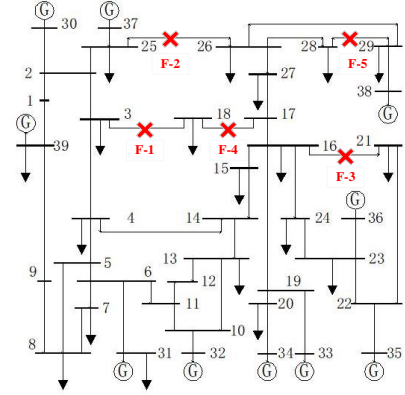

**Fig. 5.** IEEE 39-bus test system line fault diagram.

The five target domain datasets, generated by time-domain simulation using PSD-BPA, are detailed in TABLE III, where F-1, F-2, …, F-5 denote the target domain datasets corresponding to five specific failures.

TABLE III
COMPONENTS OF EACH DATASET

| Samples | F-1 | F-2 | F-3 | F-4 | F-5 |
|---|---|---|---|---|---|
| Stable samples | 968 | 635 | 733 | 715 | 948 |
| Unstable samples | 1012 | 1345 | 1247 | 1265 | 1032 |

The inputs of comparison methods are the same, and the original data is processed consistently. Raw data is generated using commercial simulation software, labeled with temporal ensembling, normalized, and expanded via LSGAN, which ensures fair evaluation of our proposed method and comparison algorithms.

*B. Performance testing of transfer learning effects*

To examine the effects of transfer leaning, performance testing before and after TL have been conducted. The test results are listed in Table IV.

TABLE IV
COMPARISON OF THE EFFECTIVENESS OF TRANSFER LEARNING

| Item | ACC | AUC | MCC | F1-score |
|---|---|---|---|---|
| Before TL | 0.7862 | 0.7553 | 0.7026 | 0.7135 |
| After TL | 0.9847 | 0.9983 | 0.9661 | 0.9847 |

From reviewing Table IV above, the various evaluation metrics show significant improvement after the application of transfer learning compared to before, with an approximate accuracy difference of 0.2. This discrepancy originates from the STVSA model's struggle to accurately discern datasets from distinct domains without the aid of transfer learning. However, this issue can be effectively mitigated via transfer learning.

*C. Performance testing of temporal ensembling*

To properly evaluate the efficacy of labeling samples via temporal ensembling, we have conducted comparative tests between the utilized approach and other semi-supervised clustering methods. Temporal ensembling strategically harnesses a minor subset of samples with pre-established labels as instructive, and orchestrates clustering by integrating them into the objective function of temporal ensembling, thereby deriving labels for the entire dataset.

This study employs the temporal ensembling, semi-supervised fuzzy c-means clustering (SFCM), k-means clustering algorithm (K-means), and the engineering criteria outlined in [26] to discern four distinct labeled datasets. In this experiment, the SCs of these four labeled datasets are computed and utilized as performance metrics. The summarized performance results can be found in TABLE V below.

TABLE V
COMPARISON OF CLUSTERING PERFORMANCE

| Methods | Temporal ensembling | SFCM | K-means | Engineering criterion |
|---|---|---|---|---|
| SC | 0.5528 | 0.5002 | 0.3854 | 0.2686 |

As shown in TABLE V, the SC associated with temporal ensembling surpasses that of SFCM, COP-k-means, and the engineering criterion. This finding demonstrates that the temporal ensembling, in comparison to other methods, is more adept at extracting the underlying patterns within the dataset and yields a more reliable labeled dataset.

*D. Performance testing of data augmentation*

In order to assess the impact of LSGAN-based data augmentation, we have performed performance tests both before and after DA. By leveraging data augmentation, the initial 1980 samples are expanded to 8000 in the target domain dataset. The test results are presented in Table VI.

TABLE VI
COMPARISON OF LSGAN EFFECTIVENESS

| Item | ACC | AUC | MCC | F1-score |
|---|---|---|---|---|
| Before DA | 0.9236 | 0.9301 | 0.8981 | 0.9001 |
| After DA | 0.9847 | 0.9983 | 0.9661 | 0.9847 |

From Table VI, it is evident that data augmentation significantly contributes to the enhancement of the generalization ability of the presented STVSA model. It should be noted that the size of the target domain dataset significantly influences the performance of the proposed method. Limited samples can lead to overfitting in deep learning models. Data augmentation enhances data diversity, but excessive augmentation can introduce significant noise and unnecessary complexity, deteriorating generalization capability. Therefore, achieving the appropriate balance in dataset size is crucial for enhancing the performance of the STVSA mode.

*E. Robustness test under a noisy environment*

In order to scrutinize the robustness of the proposed method within a noisy environment, we conduct noise testing under various noise intensities. Real-world applications inevitably encounter noise disruptions when the Phasor Measurement Units (PMUs) perform real-time sampling. Consequently, this paper considers PMU measurement noises at different Signal-to-Noise Ratios (SNRs). In this case, white Gaussian noises at SNRs of 30dB, 40dB, and 50dB are introduced to the F-1 testing dataset, with the test results delineated in TABLE VII.

TABLE VII
PERFORMANCE COMPARISON UNDER DIFFERENT SNRS

| SNR (dB) | ACC | AUC | MCC | F1-score |
|---|---|---|---|---|
| Noise-free | 0.9847 | 0.9983 | 0.9661 | 0.9847 |
| 30dB | 0.9781 | 0.9969 | 0.9547 | 0.9784 |
| 40dB | 0.9717 | 0.9966 | 0.9431 | 0.9715 |
| 50dB | 0.9828 | 0.9972 | 0.9671 | 0.9827 |

TABLE VII shows our method continues to demonstrate strong performance, even when white Gaussian noises with various SNRs are incorporated into the testing datasets. This illustrates the robustness of our method under noise conditions.

To validate the superiority of our proposed method over other STVSA methods in a noisy environment, we conducted a comparative analysis with 30 dB of noise in the F-1 testing dataset. The outcomes of this test are tabulated in TABLE VIII.

TABLE VIII
PERFORMANCE COMPARISON IN NOISY ENVIRONMENTS

| Method | ACC | AUC | MCC | F1-score |
|---|---|---|---|---|
| Our method | 0.9781 | 0.9969 | 0.9547 | 0.9784 |
| LSTM [22] | 0.9265 | 0.9625 | 0.9058 | 0.9236 |
| DT [19] | 0.9046 | 0.9564 | 0.8931 | 0.9013 |
| SVM [32] | 0.8264 | 0.8961 | 0.8276 | 0.8137 |

TABLE VIII reveals that our method significantly outperforms other approaches across various evaluation metrics, affirming the exceptional robustness of our proposed method amidst noisy conditions.

*F. Adaptability test to topological changes*

In this section, we have trained five deep learning models using five distinct fault datasets: F-A, F-B, F-C, F-D, and F-E. To assess the efficacy of transfer learning, each model is tested across all these datasets. Using the model's performance on its original dataset as a benchmark helps gauge the effectiveness of the transfer. The metrics Acc-F-A, Acc-F-B, ..., Acc-F-E denote the accuracy achieved by each model when applied to the datasets F-A, F-B, ..., F-E, respectively. Comprehensive results from these experiments can be seen in **Fig. 6**, where the x-axis represents the fault datasets and the y-axis displays the prediction accuracy.

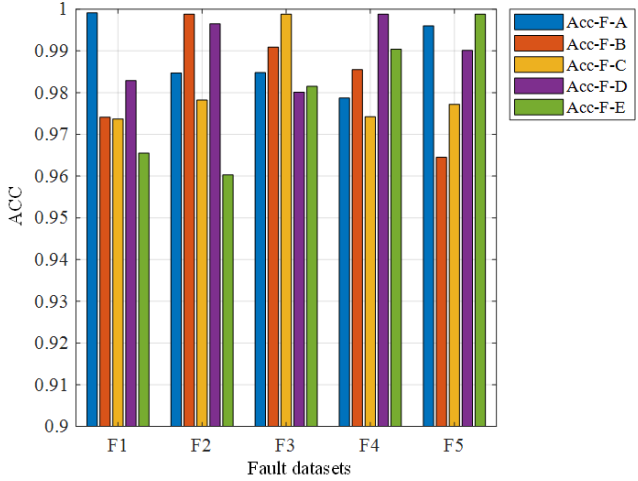

**Fig. 6.** Accuracy comparison after transfer learning under different faults.

When the source domain dataset is utilized to train the model, we observe the prediction accuracy nearing an optimal 100%. Upon transferring to each of the four alternate fault datasets, the model experiences an anticipated decline in accuracy. Nonetheless, even at its lowest point, the model preserves an accuracy exceeding 95%, substantiating its efficacy across different fault types. Hence, this method exhibits significant adaptability to topological alterations.

*G. Effects of varying lengths of observation time windows*

The selection of an appropriate OTW for a specific power system is crucial in STVSA. Specifically, choosing a smaller window size may result in quick, yet imprecise evaluations, whereas opting for a larger window size can yield precise but tardy assessments. Therefore, we test the performance of our approach against other alternatives with varied OTWs. The results from these comparative tests are presented in TABLE IX below.

TABLE IX
THE DIFFERENT PERFORMANCE INDICATORS OF DIFFERENT METHODS

| Methods | OTW | ACC | AUC | MCC | F1-score |
|---|---|---|---|---|---|
| Transformer | **0.03** | **0.9671** | **0.9935** | **0.9451** | **0.9651** |
|  | 0.06 | 0.9711 | 0.9938 | 0.9514 | 0.9739 |
|  | 0.09 | 0.9784 | 0.9956 | 0.9536 | 0.9825 |
|  | 0.12 | 0.9847 | 0.9983 | 0.9661 | 0.9847 |
| LSTM [22] | 0.03 | 0.9267 | 0.9814 | 0.8554 | 0.9259 |
|  | 0.06 | 0.9288 | 0.9828 | 0.8519 | 0.9294 |
|  | 0.09 | 0.9416 | 0.9865 | 0.8860 | 0.9429 |
|  | **0.12** | **0.9442** | **0.9884** | **0.8858** | **0.9449** |
| DT [19] | 0.03 | 0.8902 | 0.9561 | 0.8852 | 0.8911 |
|  | 0.06 | 0.8970 | 0.9636 | 0.8950 | 0.8982 |
|  | 0.09 | 0.9011 | 0.9662 | 0.9148 | 0.9029 |
|  | 0.12 | 0.9119 | 0.9743 | 0.9118 | 0.9263 |
| SVM [32] | 0.03 | 0.8562 | 0.9309 | 0.8384 | 0.8437 |
|  | 0.06 | 0.8562 | 0.9472 | 0.8439 | 0.8437 |
|  | 0.09 | 0.8562 | 0.9557 | 0.8514 | 0.8437 |
|  | 0.12 | 0.8562 | 0.9581 | 0.8552 | 0.8437 |

TABLE IX clearly demonstrates that our proposed method is far superior to all other alternatives. Particularly noteworthy is the fact that when the OTW is set to 0.03s, our method achieves an accuracy of 0.9671, surpassing the accuracy of the Transfer Learning-LSTM model with an OTW of 0.12s by 0.02. This indicates that once a large disturbance occurs, our approach can evaluate the stability status of the power system more swiftly and accurately, thereby providing further validation of its superiority. Moreover, the proposed method meets the real-time requirements for STVSA in power systems.

## VI. CONCLUSION

To overcome the challenges faced by existing machine learning-based STVSA approaches, including limited adaptability to topological changes, difficulties in labeling samples, and inefficiencies in handling small datasets, this paper introduces a method that utilizes deep transfer learning and PMU measurements. Based on the results obtained from the case studies, the following conclusions can be drawn:
1) The proposed method effectively overcomes the limitations of existing STVSA methods in adapting to network topology changes by leveraging deep transfer learning. It deeply mines the correlation between different fault datasets, showcasing its adaptability and novelty in the STVSA field.
2) The presented method manages to address the challenges associated with sample labeling and small dataset handling. This is achieved by leveraging the temporal ensembling for labeling samples and employing LSGAN to generate good synthetic samples, thereby improving the model's performance.
3) The integration of PMUs in the online assessment stage enables real-time evaluation of voltage stability. By harnessing the precision and timeliness of PMU measurements, this approach enhances the performance of STVSA, highlighting the significant role of advanced instrumentation in power system stability assessments.

4) The test results showcase the excellent performance and robustness of the proposed Transformer-based method in noisy environments. By leveraging the power of the self-attention mechanism, our approach outperforms other deep learning and shallow learning models in terms of better performance and shorter observation time.

In this study, we assume that all buses are installed with PMUs, but more sophisticated scenarios necessitate the consideration of optimal PMU placement. Future research could explore the stability assessment under partial PMU information missing scenarios. Additionally, employing real-world data would be valuable for further examining the performance of the proposed method [42, 43].